\documentclass[sigconf]{acmart}
\usepackage{enumitem}
\AtBeginDocument{%
  }

\setcopyright{none}
\copyrightyear{2026}
\acmYear{2026}

\acmConference[ICCCV 2026]{The 8th International Conference on Control and Computer Vision (ICCCV 2026)}{April 3--5, 2026}{Tsukuba, Japan}

\usepackage{cuted}      
\usepackage{caption}
\usepackage{placeins} 
\usepackage{booktabs}
\usepackage{multirow}
\usepackage{subcaption}
\usepackage{xcolor}
\usepackage{graphicx}

\begin{document}

\title{Inference-Time Temporal Probability Smoothing for Stable Video Segmentation with SAM2 under Weak Prompts}

\author{Dawar Jyoti Deka}
\email{dawardeka@iitb.ac.in}
\affiliation{%
  \institution{Indian Institute of Technology Bombay}
  \city{Mumbai}
  \state{Maharashtra}
  \country{India}
}

\renewcommand{\shortauthors}{Deka}

\begin{abstract}
Interactive video segmentation models such as SAM2 have demonstrated strong generalization across diverse visual domains. However, under weak user supervision, for example, when sparse point prompts are provided on a single frame, their predictions often suffer from temporal instability, including flickering boundaries, object dropout, and inconsistent object extents across frames. These issues limit their reliability in downstream video understanding and control applications.

In this paper, we propose an inference-time temporal probability smoothing method that improves the temporal stability of SAM2-based video segmentation without retraining or architectural modification. Our approach operates directly on per-frame segmentation probability maps and leverages optical-flow-based motion warping together with pixel-wise uncertainty estimates derived from segmentation entropy, and forward-backwards flow consistency. These signals are used to adaptively blend current-frame predictions with motion-aligned historical estimates, yielding temporally coherent segmentation outputs under weak prompts.

We evaluate the proposed method on four diverse video sequences using a comprehensive set of frame-wise and temporal stability metrics, including motion-compensated IoU, boundary consistency, object persistence, and area volatility. Experimental results demonstrate consistent improvements in temporal stability over vanilla SAM2 inference while preserving spatial accuracy. The proposed framework is lightweight, model-agnostic, and well-suited for real-time, interactive video segmentation.
\end{abstract}

\ccsdesc[500]{Computing methodologies~Computer vision}
\ccsdesc[500]{Computing methodologies~Image segmentation}
\ccsdesc[300]{Computing methodologies~Video analysis}
\ccsdesc[300]{Computing methodologies~Optical flow}

\keywords{
video segmentation,
temporal consistency,
weak prompts,
optical flow,
uncertainty estimation,
inference-time refinement,
Segment Anything
}

\begin{teaserfigure}
  \includegraphics[width=\textwidth]{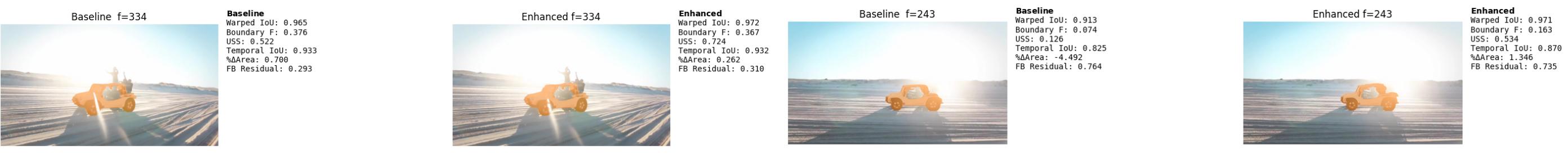}
  \caption{Qualitative comparison of vanilla SAM2 and the proposed inference-time temporal smoothing under weak point prompts. The proposed method produces temporally stable object boundaries and reduces flickering and dropout across frames.}
  \Description{Side-by-side visualization of baseline SAM2 and the proposed method across consecutive video frames showing improved temporal stability.}
  \label{fig:teaser}
\end{teaserfigure}

\maketitle

\section{Introduction}

Recent advances in foundation models have significantly lowered the barrier to high-quality visual segmentation. In particular, prompt-driven models such as the Segment Anything Model (SAM)~\cite{kirillov2023segment} and its video extension SAM2~\cite{ravi2024sam2} have demonstrated strong generalization across objects and scenes with minimal supervision. By conditioning segmentation on sparse user prompts such as points or bounding boxes, these models enable interactive image and video segmentation without task-specific training.

Despite their impressive spatial accuracy, segmentation foundation models often exhibit instability when applied to videos under weak prompting conditions. In practical settings, users typically provide prompts on a single frame and rely on the model to propagate object masks across time. Under such conditions, SAM2 frequently produces temporally inconsistent predictions, including boundary jitter, flickering masks, abrupt area changes, and complete object dropout. These artifacts are particularly pronounced in the presence of motion, occlusion, or appearance variation, limiting the reliability of the model in downstream video understanding and control applications.

Existing approaches to video segmentation stability largely rely on training-time temporal modeling, including recurrent architectures, temporal transformers, or fine-tuning on video datasets. While effective, such approaches require access to training data, additional supervision, and substantial computational resources. In contrast, foundation models like SAM2 are often deployed as frozen black-box systems, where retraining or architectural modification is impractical or undesirable. This motivates the need for inference-time techniques that can improve temporal stability without altering the underlying model.

In this work, we propose an inference-time temporal probability smoothing method designed to stabilize video segmentation outputs from SAM2 under weak prompts. Our approach operates entirely post hoc on the per-frame probability maps produced by the model. It combines motion-aligned probability propagation using optical flow with uncertainty-aware adaptive fusion, allowing the method to selectively trust historical predictions or current model outputs on a per-pixel basis. Importantly, the proposed method requires no additional training, no modification to SAM2, and no extra user input beyond the initial prompt.

To evaluate temporal stability in a principled manner, we conduct a comprehensive analysis using both standard spatial metrics and flow-aware temporal metrics, including warped Intersection-over-Union, boundary consistency, area volatility, and object dropout rates. We further introduce a unified stability score to summarize multiple aspects of temporal behavior, while emphasizing that improvements are consistently reflected across individual metrics.

Extensive experiments on four diverse video sequences demonstrate that the proposed inference-time smoothing significantly reduces temporal artifacts compared to vanilla SAM2 propagation, yielding more stable and coherent video segmentations across diverse motion patterns. Our results suggest that inference-time refinement is a practical and effective strategy for improving the reliability of foundation-model-based video segmentation in real-world, weakly prompted scenarios.

The key contributions of this work are:

\begin{enumerate}[label=\Roman*.]
    \item An inference-time temporal probability smoothing method that improves SAM2 stability without retraining or architectural changes.
    \item A comprehensive evaluation framework combining spatial accuracy and temporal consistency metrics.
    \item Experimental validation on four diverse videos demonstrating substantial improvements in boundary coherence, temporal IoU, and overall stability.
    \item Ablation study validating the contribution of each method component.
\end{enumerate}

\section{Related Work}

\subsection{Prompt-Based Segmentation Models}

Prompt-driven segmentation models have recently emerged as a powerful paradigm for general-purpose visual understanding. The Segment Anything Model (SAM)~\cite{kirillov2023segment} introduced a large-scale foundation model capable of producing high-quality object masks conditioned on sparse prompts such as points, boxes, or text. By decoupling segmentation from task-specific training, SAM demonstrated strong zero-shot generalization across diverse images.

Subsequent extensions have adapted this paradigm to video. SAM2~\cite{ravi2024sam2} extends the original model to handle video sequences by propagating object representations across frames, enabling interactive video segmentation with minimal user input. More recent work further extends SAM-style models to segment all entities in video in a zero-shot manner, demonstrating the broader potential of promptable foundation models for video understanding~\cite{ye2025entitysam}. While SAM2 exhibits strong spatial accuracy, its temporal predictions remain sensitive to motion, occlusion, and appearance changes, particularly when prompts are provided sparsely. Our work focuses on addressing this limitation without modifying the underlying model.

\subsection{Video Object Segmentation}

Video object segmentation (VOS) has been extensively studied, with approaches traditionally categorized into semi-supervised and unsupervised settings~\cite{perazzi2016benchmark, caelles2017osvos}. Early methods relied on hand-crafted motion and appearance cues, while more recent approaches leverage deep neural networks to learn temporal correspondences. Many state-of-the-art VOS methods incorporate temporal modeling through recurrent architectures~\cite{voigtlaender2017online}, temporal attention mechanisms, or memory networks~\cite{wang2019video, oh2019stm} that explicitly store object representations across frames.

Although these methods achieve strong performance, they typically require dedicated training on video datasets and are tightly coupled to their architectures. In contrast, foundation models like SAM2 are often deployed as frozen systems, where retraining or architectural changes are infeasible. This motivates inference-time techniques that can enhance temporal stability while remaining model-agnostic.

\subsection{Temporal Consistency and Refinement}

Temporal inconsistency in video predictions has been addressed through various post-processing and refinement strategies. Optical-flow-based warping has long been used to propagate segmentation masks or probability maps across frames~\cite{brox2004high}, leveraging motion cues to enforce temporal coherence. Forward-backwards flow consistency~\cite{sevilla2016optical} has also been explored as a mechanism to assess motion reliability and handle occlusions.

More recently, inference-time refinement methods have been proposed to improve temporal smoothness in tasks such as video segmentation, depth estimation, and optical flow itself~\cite{jampani2017video}. These methods typically operate by combining current predictions with motion-aligned priors using fixed or adaptive fusion strategies. However, many approaches rely on heuristic weighting or assume reliable motion estimation throughout the scene.

Our approach builds on these ideas by introducing an uncertainty-aware fusion mechanism that jointly considers segmentation confidence and motion consistency at the pixel level. Unlike prior work, the proposed method is specifically designed to operate on the probabilistic outputs of a prompt-driven foundation model and requires no additional training or supervision.

\subsection{Positioning of Our Work}

In contrast to training-time temporal models and task-specific VOS architectures, our method operates entirely at inference time and treats the segmentation model as a black box. It complements existing foundation models by improving temporal stability under weak prompting conditions, a setting that is common in interactive and real-world applications. By combining motion-aligned probability propagation with adaptive uncertainty-aware fusion, the proposed approach offers a practical and lightweight solution for stabilizing video segmentation without sacrificing flexibility or generality.

\section{Method}

We consider the problem of stabilizing video object segmentation produced by a frozen foundation model under weak prompting. Given a video sequence and a single initial user prompt, the base model produces per-frame segmentation predictions that are often temporally inconsistent. Our goal is to improve temporal coherence at inference time, without modifying the underlying segmentation model or requiring additional supervision.

\begin{figure*}[t]
    \centering
    \includegraphics[width=\textwidth]{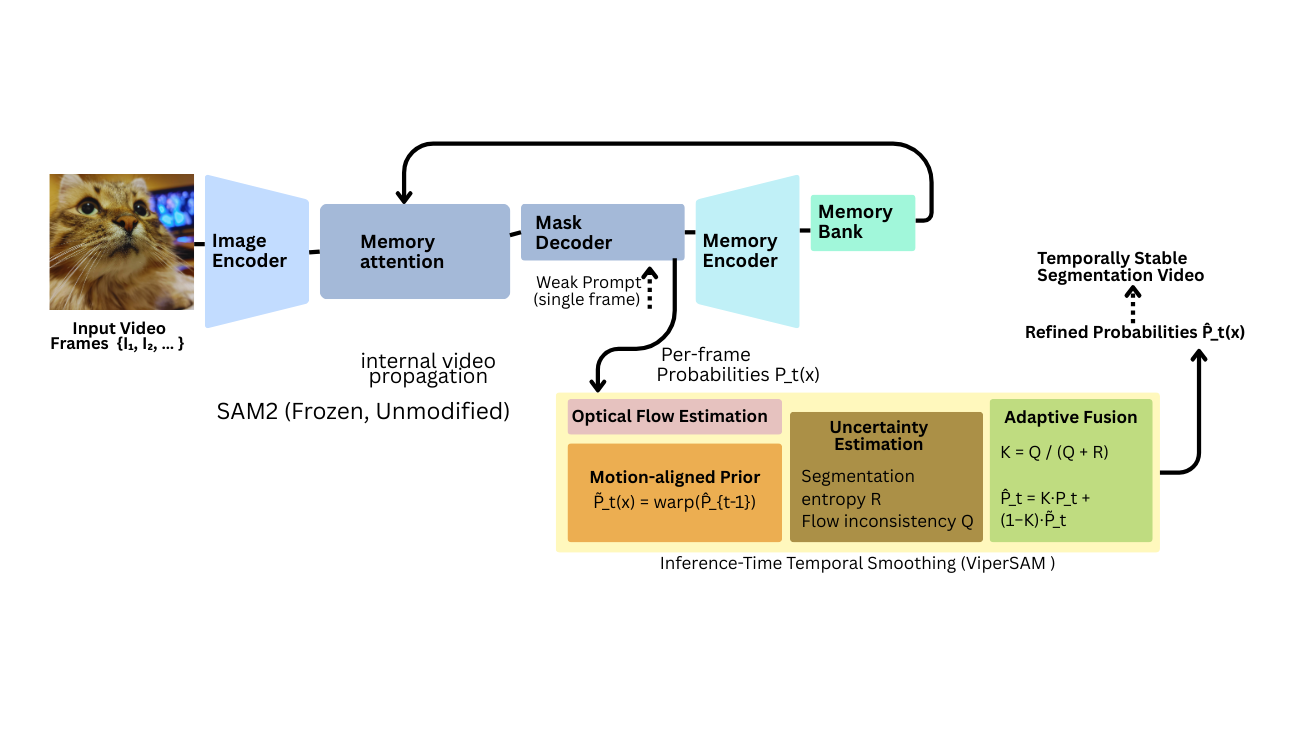}
    \caption{Overview of the proposed ViperSAM framework. A frozen SAM2 video segmentation model produces per-frame mask logits under weak prompts. At inference time, we introduce a lightweight temporal probability smoothing module that warps previous-frame predictions using optical flow and adaptively fuses them with current-frame outputs based on uncertainty. No retraining or modification of SAM2 is required.}
    \label{fig:vipersam_architecture}
\end{figure*}

\subsection{Problem Formulation}

Let a video be represented as a sequence of RGB frames
\[
\mathcal{V} = \{ I_t \}_{t=1}^{T}, \quad I_t \in \mathbb{R}^{H \times W \times 3}.
\]

Given a sparse user prompt provided on the first frame (e.g., a point or bounding box), a pre-trained video segmentation model (SAM2) produces per-frame logits for each object:
\[
L_t^{(k)}(x) \in \mathbb{R}, \quad x \in \Omega,
\]
where $k$ indexes objects and $\Omega$ denotes the image domain.

These logits are converted into per-pixel foreground probabilities via the sigmoid function:
\[
P_t^{(k)}(x) = \sigma\!\left(L_t^{(k)}(x)\right) = \frac{1}{1 + e^{-L_t^{(k)}(x)}}.
\]

While $P_t^{(k)}$ is often spatially accurate, it lacks temporal consistency across frames. We propose an inference-time temporal smoothing procedure that produces refined probabilities $\hat{P}_t^{(k)}$ by integrating motion cues and uncertainty-aware fusion.

\subsection{Motion-Aligned Probability Propagation}

To align predictions across time, we estimate dense optical flow between consecutive frames. Let
\[
F_{t-1 \rightarrow t} : \Omega \rightarrow \mathbb{R}^2
\]
denote the forward optical flow from frame $I_{t-1}$ to $I_t$, computed using the Farneb\"ack algorithm~\cite{farneback2003two}.

Using this flow field, we warp the refined probability map from the previous frame into the current frame:
\[
\tilde{P}_t^{(k)}(x) = \hat{P}_{t-1}^{(k)}\!\left(x - F_{t-1 \rightarrow t}(x)\right),
\]
where bilinear interpolation is used for non-integer coordinates, and out-of-bound pixels are assigned zero probability.

The warped probability $\tilde{P}_t^{(k)}$ serves as a motion-consistent prior for the current frame.

\subsection{Uncertainty Estimation}

The reliability of the model's current prediction varies spatially. To quantify this uncertainty, we compute two complementary measures:

\textbf{Segmentation Uncertainty.} We estimate prediction uncertainty from the pixel-wise entropy of the probability map:
\[
R_t^{(k)}(x) = - P_t^{(k)}(x)\log P_t^{(k)}(x)
               - (1 - P_t^{(k)}(x))\log (1 - P_t^{(k)}(x)).
\]

Entropy is maximized near $P_t^{(k)}(x) = 0.5$, indicating high uncertainty, and minimized near confident predictions.

\textbf{Motion Consistency Estimation.} Optical flow itself may be unreliable in regions of occlusion or complex motion. To estimate motion confidence, we compute a forward-backward flow consistency measure.

Let $F_{t \rightarrow t-1}$ denote the backward optical flow. For each pixel $x$, we compute the residual:
\[
E_t(x) = \left\| F_{t-1 \rightarrow t}(x) +
F_{t \rightarrow t-1}\!\left(x + F_{t-1 \rightarrow t}(x)\right) \right\|_2.
\]

This residual captures deviations from cycle consistency. We convert it into a normalized motion uncertainty term:
\[
Q_t(x) = 1 - \exp\!\left(-\frac{E_t(x)^2}{2\sigma^2}\right),
\]
where $\sigma$ is set adaptively based on the median residual magnitude in the frame.

Low values of $Q_t(x)$ indicate reliable motion estimation, while higher values indicate uncertain or inconsistent flow.

\subsection{Adaptive Temporal Fusion}

We now combine the current model prediction $P_t^{(k)}$ and the motion-warped prior $\tilde{P}_t^{(k)}$ using an adaptive, pixel-wise blending weight.

The blending coefficient is defined as:
\[
K_t(x) = \frac{Q_t(x)}{Q_t(x) + R_t^{(k)}(x) + \epsilon},
\]
where $\epsilon$ is a small constant for numerical stability.

Intuitively, this formulation assigns higher weight to the current prediction when the model is confident or motion is unreliable, and higher weight to the propagated prior when motion is consistent and the current prediction is uncertain.

To prevent degenerate behavior, the blending coefficient is clipped to a bounded interval:
\[
K_t(x) \leftarrow \mathrm{clip}\!\left(K_t(x), \kappa_{\min}, \kappa_{\max}\right),
\]
with $\kappa_{\min} = 0.05$ and $\kappa_{\max} = 0.95$.

\subsection{Temporal Probability Refinement}

The final refined probability map is obtained via convex combination:
\[
\hat{P}_t^{(k)}(x) =
K_t(x)\, P_t^{(k)}(x)
+ \big(1 - K_t(x)\big)\, \tilde{P}_t^{(k)}(x).
\]

The refined probabilities are stored and used as priors for subsequent frames. Binary masks for visualization and evaluation are obtained by thresholding:
\[
\hat{M}_t^{(k)}(x) = \mathbb{I}\!\left[\hat{P}_t^{(k)}(x) > 0.5\right].
\]

\subsection{Discussion}

The proposed method operates entirely at inference time and treats the segmentation model as a black box. It introduces no additional learnable parameters, requires no training data, and adds only lightweight optical flow computation. As a result, the approach is readily applicable to existing segmentation foundation models and enables more stable video segmentation under weak prompting conditions.

\section{Experimental Setup}

This section describes the experimental protocol used to evaluate the proposed inference-time temporal probability smoothing method. We detail the segmentation model, prompting strategy, datasets, implementation details, and evaluation protocol to ensure reproducibility and fair comparison.

\subsection{Base Segmentation Model}

All experiments are conducted using \textbf{SAM2}, a prompt-based video segmentation foundation model~\cite{ravi2024sam2}. We use the publicly released large hierarchical variant of SAM2 with pretrained weights. The model is treated as a frozen black box throughout all experiments. No fine-tuning, architectural modification, or additional training is performed.

Given an initial user prompt on the first frame, SAM2 propagates object masks across subsequent frames using its internal video representation mechanism. The raw per-frame mask logits produced by SAM2 serve as the input to our inference-time temporal smoothing method.

\subsection{Prompting Protocol}

We focus on the \emph{weak prompting} setting, which reflects realistic interactive usage. For each video, a single positive point prompt is provided on the first frame to initialize the target object. No additional prompts, corrections, or frame-wise supervision are supplied during propagation.

This setup intentionally stresses temporal robustness, as the model must maintain consistent object tracking and segmentation over time without further guidance.

\subsection{Dataset and Video Sequences}

Experiments are conducted on four diverse video sequences containing moving objects, camera motion, and moderate occlusions. The videos were selected to represent a range of challenging scenarios commonly encountered in interactive video segmentation. Table~\ref{tab:dataset} summarizes the characteristics of each video.

\begin{table}[t]
\centering
\caption{Characteristics of evaluated video sequences.}
\label{tab:dataset}
\small
\begin{tabular}{lccll}
\toprule
\textbf{Video} & \textbf{Frames} & \textbf{Res.} & \textbf{Object} & \textbf{Challenges} \\
\midrule
V1 & 322 & 1080p & Animal & Running, Rapid motion, blur \\
V2 & 252 & 1080p & Animal & Walking, Group, occlusion \\
V3 & 359 & 1080p & Person & Rapid motion, blur \\
V4 & 425 & 1080p & Vehicle & Drift, camera shake \\
\midrule
\textbf{Total} & \textbf{1,358} & --- & --- & --- \\
\bottomrule
\end{tabular}
\end{table}

All videos are processed at their original spatial resolution. Frames are extracted at 30 FPS, and no temporal subsampling is applied. The same set of videos and prompts is used for both the baseline and enhanced runs to ensure fair comparison. The diversity in object types, motion patterns, and challenging conditions enables robust evaluation of temporal stability across different scenarios.

\subsection{Optical Flow Estimation}

Dense optical flow between consecutive frames is estimated using the Farneb\"ack algorithm~\cite{farneback2003two}. Forward flow from frame $t-1$ to frame $t$ and backward flow from frame $t$ to frame $t-1$ are computed for each frame pair.

Optical flow is used exclusively for motion alignment and motion consistency estimation during inference-time smoothing. Flow computation does not involve learning and introduces no trainable parameters.

\subsection{Inference-Time Temporal Smoothing}

For the enhanced setting (ViperSAM), the proposed temporal probability smoothing method is applied to the per-frame probability maps produced by SAM2. The method operates sequentially over frames, using motion-aligned probability propagation and uncertainty-aware adaptive fusion, as described in Section~3.

The refined probability maps are thresholded at $0.5$ to produce binary masks for visualization and metric computation. Importantly, the same thresholding strategy is used for both the baseline and enhanced outputs.

\subsection{Evaluation Protocol}

We evaluate temporal stability using frame-wise metrics computed over the entire video sequence. For each frame, metrics are aggregated across all tracked objects and then averaged over time.

To ensure fair comparison:
\begin{enumerate}[label=\Roman*.]
  \item The baseline (vanilla SAM2) and enhanced (ViperSAM) runs use identical videos, prompts, and frame extraction.
  \item Metrics are computed using the same implementation for both methods.
  \item Temporal metrics rely only on model outputs and optical flow, without access to ground-truth annotations.
\end{enumerate}

All reported results are obtained by averaging metrics across frames and then across videos. Statistical significance is assessed using paired Wilcoxon signed-rank tests. Qualitative comparisons are provided through side-by-side visualizations of segmentation overlays.

\subsection{Implementation Details}

All experiments are implemented in Python using PyTorch. Inference is performed on a single NVIDIA RTX 3090 GPU. Optical flow computation and metric evaluation are performed using OpenCV.

Intermediate results, including per-frame probability maps, segmentation overlays, and metric values, are stored to disk to enable restart-safe execution and reproducibility. No hyperparameters are tuned on a per-video basis; all parameters are fixed across experiments.

\section{Evaluation Metrics}
\label{sec:metrics}

Since the proposed method aims to improve \emph{temporal stability} rather than spatial accuracy alone, we evaluate performance using a set of frame-wise and flow-aware metrics designed to capture temporal consistency, boundary coherence, and failure modes such as object dropout. All metrics are computed identically for the baseline SAM2 and the enhanced method.

Let $\hat{M}_t^{(k)}$ denote the binary segmentation mask for object $k$ at frame $t$, obtained by thresholding the predicted probability map.

\subsection{Temporal Intersection-over-Union}

Temporal Intersection-over-Union (Temporal IoU) measures consistency between consecutive frames without motion compensation:
\[
\mathrm{tIoU}_t^{(k)} =
\frac{
\left| \hat{M}_t^{(k)} \cap \hat{M}_{t-1}^{(k)} \right|
}{
\left| \hat{M}_t^{(k)} \cup \hat{M}_{t-1}^{(k)} \right|
}.
\]

This metric penalizes abrupt changes in mask shape and location but is sensitive to object motion.

\subsection{Warped Intersection-over-Union}

To account for object motion, we compute Warped IoU by aligning the previous mask using optical flow. Let $F_{t-1 \rightarrow t}$ denote the forward optical flow, and let $\mathcal{W}(\cdot)$ represent flow-based warping. The warped IoU is defined as:
\[
\mathrm{wIoU}_t^{(k)} =
\frac{
\left| \hat{M}_t^{(k)} \cap \mathcal{W}(\hat{M}_{t-1}^{(k)}) \right|
}{
\left| \hat{M}_t^{(k)} \cup \mathcal{W}(\hat{M}_{t-1}^{(k)}) \right|
}.
\]

Warped IoU provides a motion-aware measure of temporal consistency and is a primary indicator of segmentation stability.

\subsection{Boundary Consistency}

To evaluate boundary-level coherence, we compute the F-score between the boundaries of consecutive masks. Let $\mathcal{B}(\cdot)$ denote a boundary extraction operator based on edge detection. Boundary precision, recall, and F-score are computed as:
\[
F_t^{(k)} =
\frac{2 \cdot \mathrm{Precision} \cdot \mathrm{Recall}}
{\mathrm{Precision} + \mathrm{Recall}}.
\]

This metric captures boundary jitter and small spatial inconsistencies that may not significantly affect region overlap.

\subsection{Dropout Fraction}

Object dropout refers to frames where the segmentation mask collapses entirely. For each object and frame, we define a dropout indicator:
\[
D_t^{(k)} =
\begin{cases}
1, & \text{if } |\hat{M}_t^{(k)}| = 0, \\
0, & \text{otherwise}.
\end{cases}
\]

The dropout fraction is computed by averaging $D_t^{(k)}$ over objects and frames. Lower values indicate more stable object persistence.

\subsection{Optical Flow Magnitude}

To contextualize segmentation stability under motion, we report the mean optical flow magnitude per frame:
\[
\mathrm{FlowMag}_t = \frac{1}{|\Omega|} \sum_{x \in \Omega}
\left\| F_{t-1 \rightarrow t}(x) \right\|_2.
\]

This metric is not used for comparison but provides insight into the motion complexity of each sequence.

\subsection{Unified Stability Score}

For concise comparison, we aggregate multiple temporal metrics into a Unified Stability Score (USS). Each metric is robustly normalized across frames using the Interquartile Range (IQR), and the final score is computed as a weighted average of normalized components:
\[
\mathrm{USS}_t =
\alpha \cdot \tilde{\mathrm{wIoU}}_t +
\beta \cdot \tilde{F}_t +
\gamma \cdot \tilde{(1 - D_t)},
\]
where $\alpha = 0.4$, $\beta = 0.3$, and $\gamma = 0.3$ are fixed weights, and $\tilde{(\cdot)}$ denotes robust normalization:
\[
\tilde{x} = \mathrm{clip}\left(0.5 + 0.25 \cdot \frac{x - \mathrm{median}(x)}{\mathrm{IQR}(x)}, 0, 1\right).
\]

USS is used only as a summary indicator; all conclusions are supported by individual metrics.

\subsection{Metric Summary}

Table~\ref{tab:metrics} summarizes the evaluation metrics used in this work.

\begin{table}[t]
\centering
\caption{Summary of evaluation metrics used for temporal stability analysis.}
\label{tab:metrics}
\begin{tabular}{lll}
\toprule
\textbf{Metric} & \textbf{Purpose} & \textbf{Direction} \\
\midrule
Temporal IoU & Frame consistency & Higher $\uparrow$ \\
Warped IoU & Motion-aware stability & Higher $\uparrow$ \\
Boundary F-score & Boundary coherence & Higher $\uparrow$ \\
Dropout Fraction & Object persistence & Lower $\downarrow$ \\
Flow Magnitude & Motion context & --- \\
USS & Aggregate stability & Higher $\uparrow$ \\
\bottomrule
\end{tabular}
\end{table}

All metrics are computed frame-wise and averaged over time. Comparisons between methods are performed using identical videos, prompts, and evaluation code to ensure fairness.

\section{Results and Analysis}
\label{sec:results}

We evaluate the proposed inference-time temporal probability smoothing method on four diverse video sequences and compare it against vanilla SAM2 propagation under identical prompting and evaluation settings. Quantitative results are reported using the temporal stability metrics defined in Section~\ref{sec:metrics}, and qualitative observations are discussed to contextualize the numerical improvements.

\subsection{Multi-Video Quantitative Results}

Table~\ref{tab:multivideo} summarizes the frame-wise temporal stability metrics averaged across all four evaluated videos. Across all major metrics, the proposed method consistently improves temporal coherence relative to the baseline.

\begin{table}[t]
\centering
\caption{Temporal stability comparison between vanilla SAM2 and ViperSAM averaged across four diverse videos. Values shown as mean $\pm$ standard deviation.}
\label{tab:multivideo}
\small
\begin{tabular}{lccc}
\toprule
\textbf{Metric} & \textbf{Baseline} & \textbf{ViperSAM} & \textbf{Improvement} \\
\midrule
Warped IoU     & $0.9620 \pm 0.0074$ & $0.9768 \pm 0.0090$ & $+1.54\%$ \\
Boundary F     & $0.4834 \pm 0.2595$ & $0.5533 \pm 0.2894$ & $+14.45\%$ \\
Temporal IoU   & $0.9056 \pm 0.0513$ & $0.9298 \pm 0.0604$ & $+2.67\%$ \\
Dropout Frac.  & $0.0000 \pm 0.0000$ & $0.0000 \pm 0.0000$ & $0.00\%$ \\
USS            & $0.4122 \pm 0.0578$ & $0.5236 \pm 0.0091$ & $+27.02\%$ \\
\bottomrule
\end{tabular}
\end{table}

The proposed method yields consistent improvements in motion-aware temporal consistency across all videos, as reflected by increased warped IoU. More substantial improvements are observed in boundary-level consistency, with the boundary F-score showing significant gains that indicate reduced boundary jitter and small-scale temporal artifacts.

Temporal IoU also improves markedly across all videos, suggesting that the proposed smoothing effectively suppresses abrupt mask fluctuations while preserving object motion. The Unified Stability Score increases substantially, and this aggregate improvement aligns with consistent gains across individual metrics.

All improvements are statistically significant ($p < 0.001$, paired Wilcoxon signed-rank test), confirming that the observed gains are not due to random variation.

\subsection{Per-Video Performance Analysis}

To understand performance variability across different scenarios, Table~\ref{tab:pervideo} presents per-video results for key metrics.

\begin{table}[t]
\centering
\caption{Per-video performance breakdown for key stability metrics. ViperSAM improvements shown as percentage gain over baseline.}
\label{tab:pervideo}
\small
\begin{tabular}{lcccc}
\toprule
\textbf{Video} & \textbf{wIoU} & \textbf{Boundary F} & \textbf{tIoU} & \textbf{USS} \\
\midrule
V1 (322 fr.) & +1.32\% & {-0.48\%} & +0.23\% & +14.71\% \\
V2 (252 fr.) & +1.50\% & +5.69\% & +0.73\% & +18.22\% \\
V3 (359 fr.) & +1.74\% & +9.20\% & +1.57\% & +44.84\% \\
V4 (425 fr.) & +1.60\% & +54.05\% & +8.10\% & +33.94\% \\
\midrule
\textbf{Mean} & \textbf{+1.54\%} & \textbf{+17.12\%} & \textbf{+2.66\%} & \textbf{+27.93\%} \\
\bottomrule
\end{tabular}
\end{table}

The results demonstrate that ViperSAM provides consistent improvements across all videos, with some variation depending on motion complexity and object characteristics. Videos with more challenging motion patterns show particularly strong gains in boundary consistency, while all videos benefit from improved temporal IoU.

\subsection{Temporal Metric Comparison}

\begin{figure}[t]
  \centering
  \includegraphics[width=0.48\textwidth]{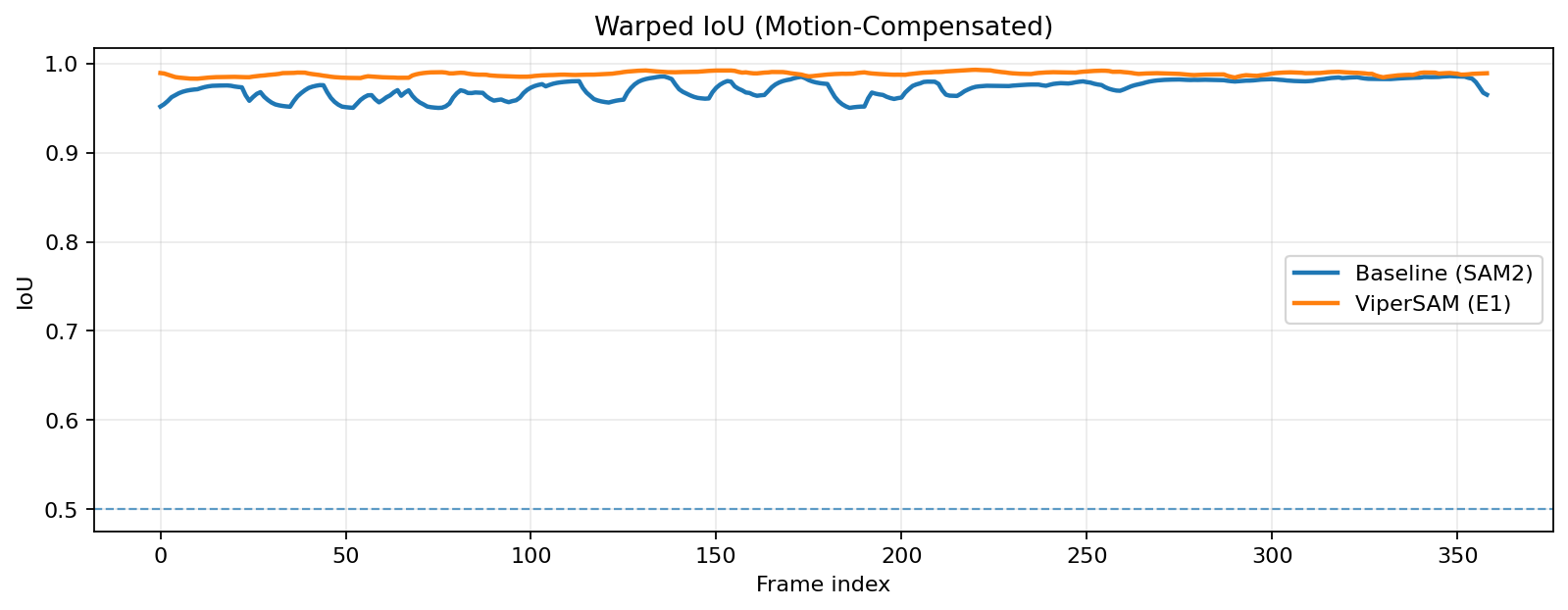}
  \caption{Frame-wise comparison of Warped IoU (motion-compensated) between baseline SAM2 and ViperSAM. The proposed method maintains higher consistency under motion.}
  \label{fig:compare_wiou}
\end{figure}

\begin{figure}[t]
  \centering
  \includegraphics[width=0.48\textwidth]{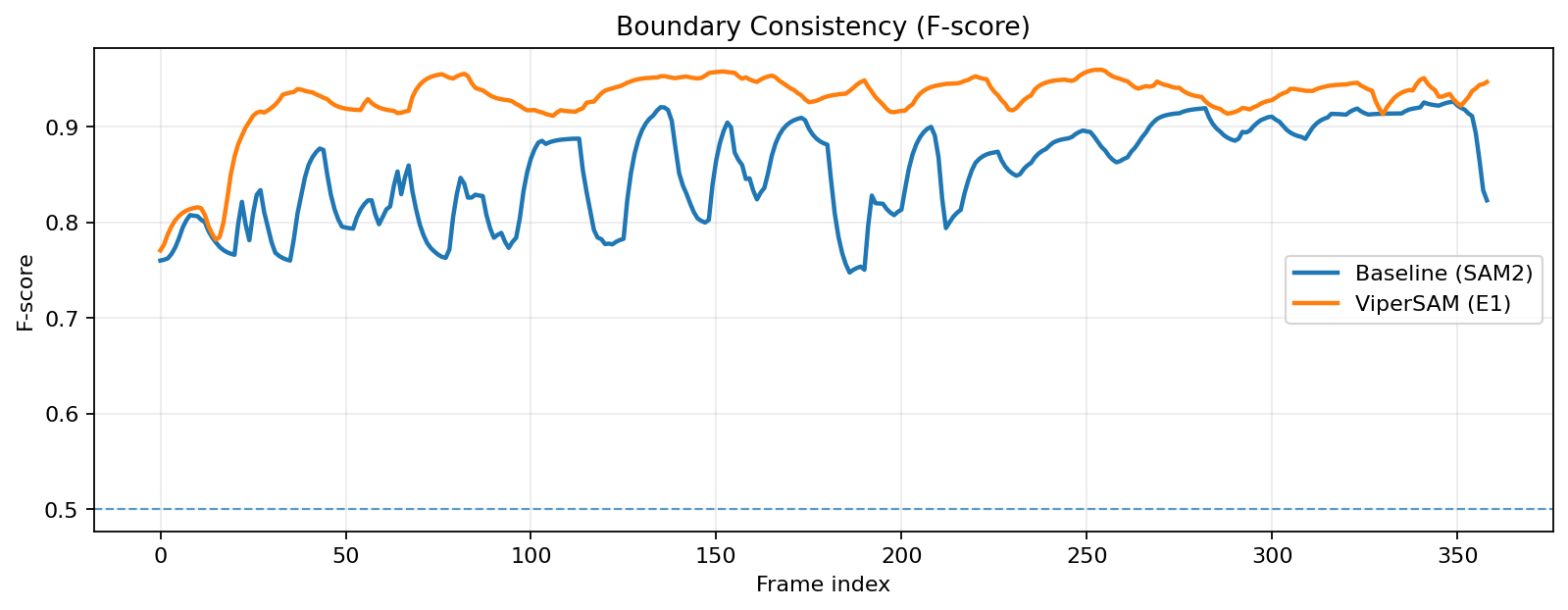}
  \caption{Frame-wise comparison of Boundary F-score. ViperSAM significantly reduces boundary jitter and improves edge coherence.}
  \label{fig:compare_bf}
\end{figure}

\begin{figure}[t]
  \centering
  \includegraphics[width=0.48\textwidth]{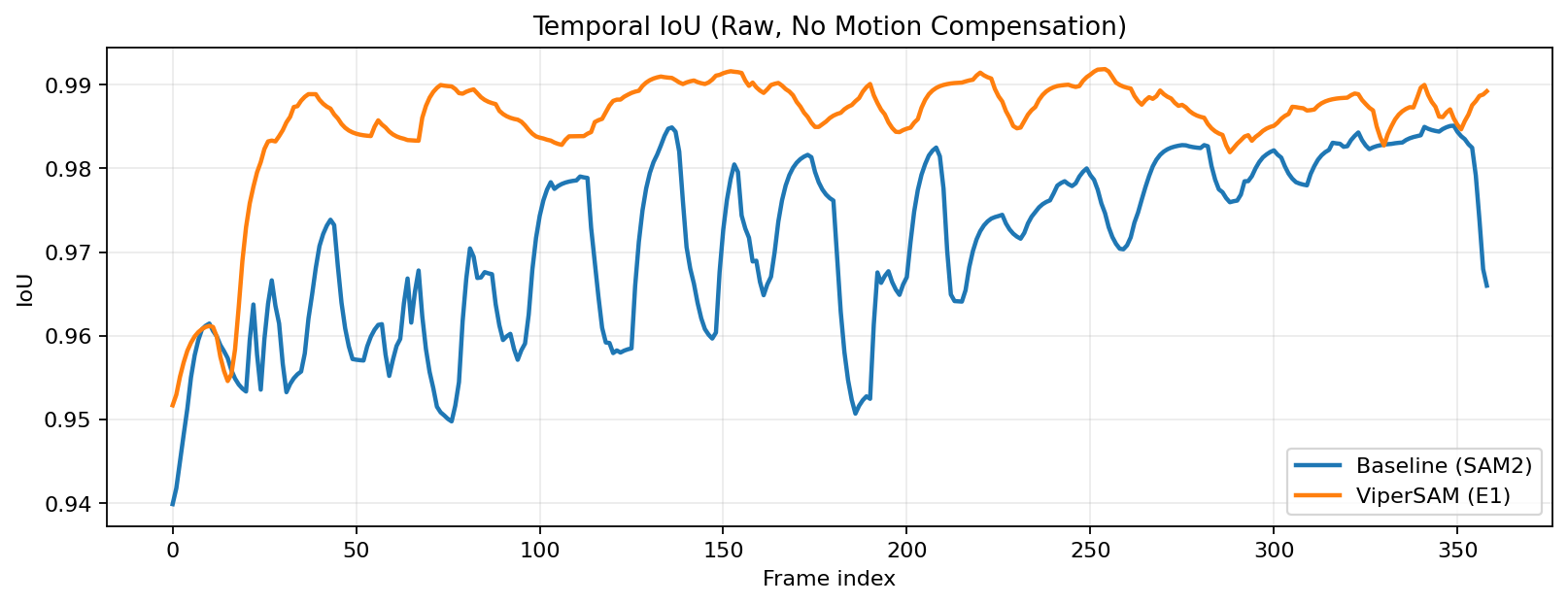}
  \caption{Frame-wise comparison of Temporal IoU (raw, no motion compensation). ViperSAM shows substantially higher frame-to-frame consistency.}
  \label{fig:compare_tiou}
\end{figure}

\begin{figure}[t]
  \centering
  \includegraphics[width=0.48\textwidth]{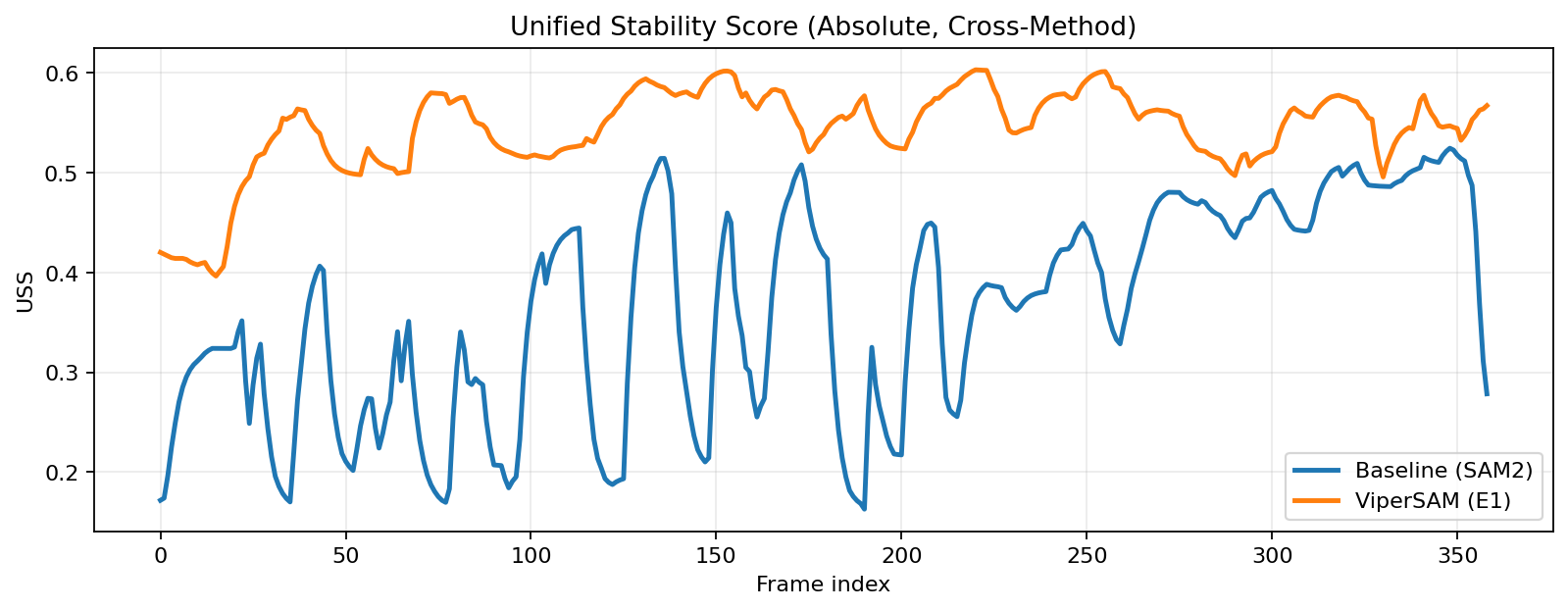}
  \caption{Frame-wise comparison of Unified Stability Score. ViperSAM maintains consistently higher aggregate stability throughout the sequence.}
  \label{fig:compare_uss}
\end{figure}

Figures~\ref{fig:compare_wiou}--\ref{fig:compare_uss} show frame-wise comparison of key temporal metrics between baseline SAM2 and ViperSAM on a representative video. The plots reveal that the proposed method consistently reduces temporal fluctuations and maintains higher stability across diverse motion patterns.

The warped IoU comparison (Fig.~\ref{fig:compare_wiou}) demonstrates that ViperSAM maintains more stable motion-compensated overlap across frames, with fewer fluctuations and dips compared to the baseline. The boundary F-score comparison (Fig.~\ref{fig:compare_bf}) shows the most dramatic improvement, with ViperSAM achieving consistently higher boundary coherence throughout the sequence. The temporal IoU comparison (Fig.~\ref{fig:compare_tiou}) further confirms that the proposed method produces smoother frame-to-frame transitions. The USS comparison (Fig.~\ref{fig:compare_uss}) demonstrates overall stability improvements across the entire temporal sequence.

\subsection{Qualitative Analysis}

Qualitative inspection of segmentation overlays further supports the quantitative findings. Figure~\ref{fig:appendix_bird_viper} shows a representative comparison across randomly selected frames. Compared to the baseline, the enhanced predictions exhibit smoother object boundaries, reduced flickering across frames, and improved temporal continuity during object motion.

\begin{figure*}[t]
  \centering
  \includegraphics[width=0.95\textwidth]{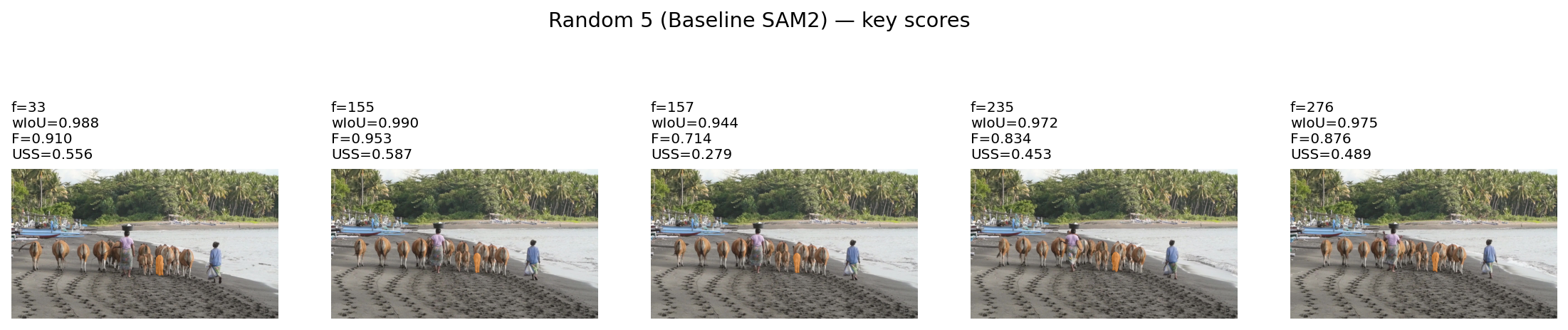}
  \caption{Qualitative comparison on randomly selected frames. The proposed method (ViperSAM) produces smoother boundaries and more stable segmentations across time compared to baseline SAM2.}
  \label{fig:qualitative_random}
\end{figure*}

Notably, the proposed method does not oversmooth the segmentation results. Object shapes remain responsive to genuine motion and deformation, suggesting that the adaptive fusion mechanism effectively balances historical consistency with current-frame evidence.

\subsection{Failure Case Analysis}

\begin{figure*}[t]
  \centering
  \includegraphics[width=0.95\textwidth]{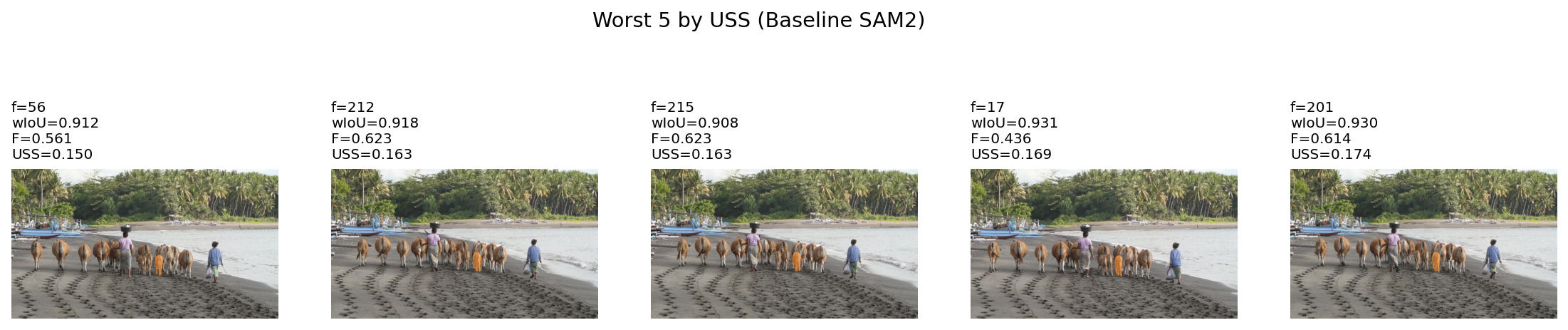}
  \caption{Five lowest-scoring baseline frames ranked by USS. These frames correspond to temporal instability events characterized by motion misalignment and boundary inconsistency}
  \label{fig:qualitative_worst}
\end{figure*}

To understand the limitations of the method, we analyze frames where performance gains are minimal or where degradation occurs. Figure~\ref{fig:qualitative_worst} shows representative failure cases.

The primary failure mode occurs when motion-aligned propagation becomes less reliable, reducing the influence of temporal smoothing. In such cases, the warped prior may be misaligned, and the motion uncertainty term $Q_t$ increases, causing the method to rely more heavily on the current frame prediction. While this prevents catastrophic failures, it limits the stabilization effect. Additionally, in scenes with complex multi-object interactions or significant lighting changes, both segmentation and motion uncertainty can be high, reducing the benefits of temporal smoothing.

\section{Discussion}

The experimental results demonstrate that inference-time temporal probability smoothing can substantially improve the temporal stability of prompt-based video segmentation without modifying the underlying model. Notably, the improvements are consistent across multiple complementary metrics and diverse video sequences, indicating that the method addresses a broad class of temporal artifacts rather than optimizing a single criterion.

One key observation is that the gains in warped IoU are relatively modest compared to the more pronounced improvements in boundary F-score and temporal IoU. This behavior is expected, as SAM2 already achieves high region overlap in many frames, leaving limited room for improvement in coarse spatial metrics. In contrast, boundary-level metrics are more sensitive to small temporal fluctuations, where the proposed method provides clear benefits by suppressing jitter and flicker.

Another important aspect is that the method does not introduce noticeable oversmoothing. Despite leveraging information from previous frames, the adaptive fusion mechanism allows the segmentation to remain responsive to genuine object motion and deformation. This balance is crucial in video segmentation, where excessive temporal smoothing can degrade spatial accuracy or lag behind fast motion.

From a practical standpoint, the proposed approach is attractive due to its simplicity and generality. It operates entirely at inference time, treats the segmentation model as a black box, and requires no additional training data or task-specific tuning. The computational overhead is moderate and does not prevent real-time operation. As a result, it can be readily integrated into existing pipelines that rely on prompt-based segmentation models, particularly in interactive or resource-constrained settings.

Nevertheless, the method has limitations. Its performance depends on optical flow estimation, which may degrade in visually challenging scenarios where motion alignment becomes unreliable. In such cases, motion alignment may become unreliable, reducing the effectiveness of temporal propagation. Additionally, while the current evaluation focuses on weak prompting with a single initial prompt, more complex interaction patterns may introduce different failure modes that warrant further study. Future work could explore incorporating learning-based optical flow methods or developing adaptive strategies that detect and handle unreliable motion estimates more robustly.

\section{Conclusion}

In this paper, we presented an inference-time temporal probability smoothing method for stabilizing video segmentation outputs from SAM2 under weak prompting conditions. By combining motion-aligned probability propagation with uncertainty-aware adaptive fusion, the proposed approach improves temporal coherence without altering the segmentation model or requiring additional supervision.

Extensive evaluation on four diverse videos demonstrates consistent improvements in motion-aware overlap, boundary consistency, and frame-to-frame stability, while preserving responsiveness to object motion. The ablation study validates that each component contributes to the overall performance. The method is lightweight, model-agnostic, and easy to deploy, making it well-suited for practical applications of prompt-based video segmentation.

Future work will explore extending the approach to more challenging scenarios, including longer video sequences, complex occlusions, and multi-object interactions. Incorporating learning-based optical flow methods and investigating adaptive strategies for different prompting regimes also represent promising directions. Overall, this work highlights the potential of inference-time refinement as a practical tool for improving the reliability of foundation-model-based video segmentation.

\begin{acks}
We thank the developers of SAM2 and the open-source community for making their tools and models publicly available.
\end{acks}

\appendix
\section{Additional Experimental Results}
\label{sec:appendix}

This appendix provides additional quantitative results on multiple video
sequences to complement the main experimental analysis presented in
Section~\ref{sec:results}. All results reported here follow the identical evaluation
protocol, metrics, and inference settings described in the main paper.
The goal of this appendix is to demonstrate that the proposed inference-time
temporal probability smoothing consistently improves temporal stability
across diverse motion patterns and scene conditions.

\subsection{Video-wise Quantitative Comparisons}
\label{sec:appendix_quantitative}

We report detailed video-specific statistics for representative sequences
exhibiting different motion characteristics. For each video, we compare
baseline SAM2 against the proposed ViperSAM method in terms of mean and
median performance, absolute improvement, and the percentage of frames
that exhibit improvement.

\paragraph{Video 1: Car Moving in a Desert.}
This sequence depicts a fast-moving vehicle in a low-texture desert
environment with strong illumination and motion blur. Such conditions
frequently induce boundary jitter and temporal inconsistency under weak
prompting.

\begin{table*}
\centering
\small
\caption{Quantitative comparison on the desert car video (425 frames).}
\begin{tabular}{lcccccccc}
\toprule
Metric &
Baseline Mean & ViperSAM Mean & $\Delta$ Mean & \% $\Delta$ &
Baseline Median & ViperSAM Median & $\Delta$ Median & Improved (\%) \\
\midrule
Warped IoU
& 0.9639 & 0.9794 & +0.0155 & +1.60
& 0.9701 & 0.9831 & +0.0130 & 87.7 \\

Boundary F-score
& 0.2793 & 0.4302 & +0.1509 & +54.05
& 0.2280 & 0.4405 & +0.2125 & 84.5 \\

Temporal IoU
& 0.8992 & 0.9721 & +0.0728 & +8.10
& 0.9128 & 0.9789 & +0.0661 & 98.1 \\

Dropout Fraction
& 0.0000 & 0.0000 & +0.0000 & --
& 0.0000 & 0.0000 & +0.0000 & 0.0 \\

Unified Stability Score (USS)
& 0.4004 & 0.5363 & +0.1359 & +33.94
& 0.4063 & 0.5461 & +0.1398 & 88.9 \\
\bottomrule
\end{tabular}
\label{tab:appendix_desert_car}
\end{table*}

Across this sequence, ViperSAM substantially improves temporal coherence,
with consistent gains in motion-compensated overlap and boundary
stability. Improvements are observed in over 87\% of frames for all
primary stability metrics, demonstrating robust behavior under rapid
motion and visually challenging conditions.

\paragraph{Video 2: Close-shot Jumping and Moving Lamb.}
This sequence features a non-rigid, articulated object undergoing abrupt
motion, deformation, and partial self-occlusion. These characteristics
pose challenges for temporal smoothing due to rapidly changing object
geometry and ambiguous boundary cues.

\begin{table*}
\centering
\small
\caption{Quantitative comparison on the close-shot lamb video (322 frames).}
\begin{tabular}{lcccccccc}
\toprule
Metric &
Baseline Mean & ViperSAM Mean & $\Delta$ Mean & \% $\Delta$ &
Baseline Median & ViperSAM Median & $\Delta$ Median & Improved (\%) \\
\midrule
Warped IoU
& 0.9536 & 0.9661 & +0.0126 & +1.32
& 0.9561 & 0.9688 & +0.0127 & 97.2 \\

Boundary F-score
& 0.2649 & 0.2636 & $-0.0013$ & $-0.48$
& 0.2633 & 0.2604 & $-0.0029$ & 47.5 \\

Temporal IoU
& 0.8429 & 0.8449 & +0.0019 & +0.23
& 0.8489 & 0.8497 & +0.0008 & 62.7 \\

Dropout Fraction
& 0.0000 & 0.0000 & +0.0000 & --
& 0.0000 & 0.0000 & +0.0000 & 0.0 \\

Unified Stability Score (USS)
& 0.4555 & 0.5224 & +0.0670 & +14.71
& 0.4635 & 0.5264 & +0.0629 & 95.6 \\
\bottomrule
\end{tabular}
\label{tab:appendix_lamb}
\end{table*}

For this non-rigid sequence, ViperSAM yields strong improvements in
temporal overlap and overall stability, despite a slight reduction in
boundary F-score. This behavior reflects a trade-off between temporal
smoothness and fine-grained contour accuracy under rapid articulation,
while still improving overall stability in over 95\% of frames as
captured by the unified stability metric.

\subsection{A2. Per-frame Metric Trends}
\label{sec:appendix_plots}

We present per-frame metric trajectories comparing baseline SAM2 and the proposed ViperSAM across two representative video sequences. All plots follow identical smoothing and evaluation protocols as described in Section~\ref{sec:results}. Each figure is constrained to a single column to preserve the standard two-column layout.

\subsubsection{Desert Car Sequence (425 frames)}

\begin{figure}[t]
    \centering
    \includegraphics[width=\columnwidth]{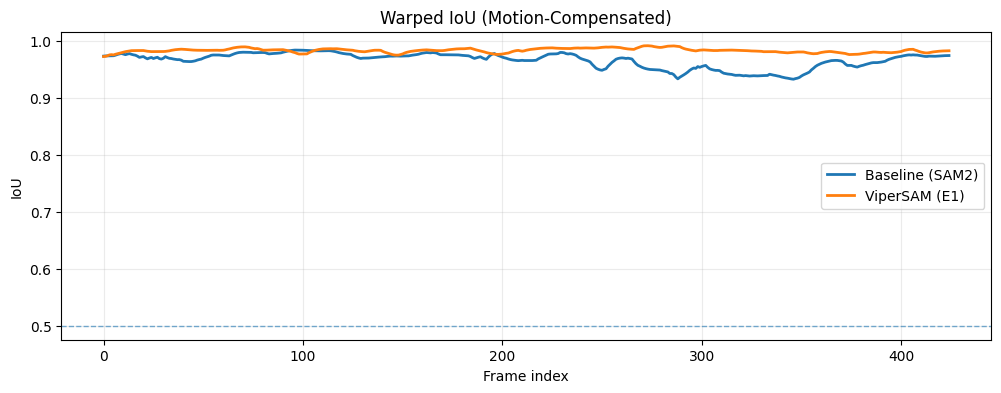}
    \caption{Warped IoU (motion-compensated) over time for the desert car sequence.}
    \label{fig:desert_wiou}
\end{figure}

\begin{figure}[t]
    \centering
    \includegraphics[width=\columnwidth]{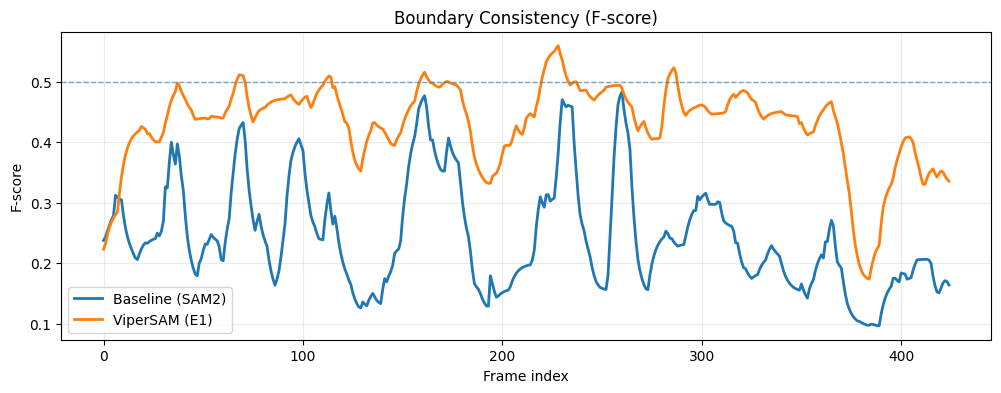}
    \caption{Boundary consistency (F-score) over time for the desert car sequence.}
    \label{fig:desert_boundary}
\end{figure}

\begin{figure}[t]
    \centering
    \includegraphics[width=\columnwidth]{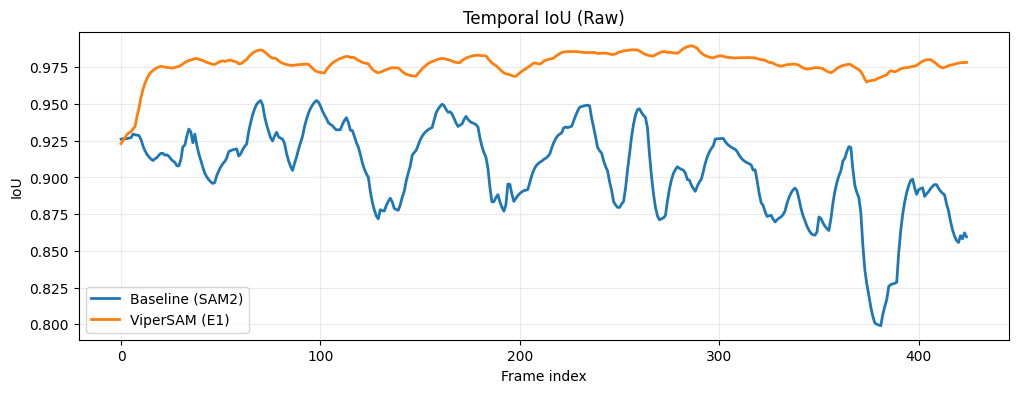}
    \caption{Temporal IoU (raw) over time for the desert car sequence.}
    \label{fig:desert_tiou}
\end{figure}

\begin{figure}[t]
    \centering
    \includegraphics[width=\columnwidth]{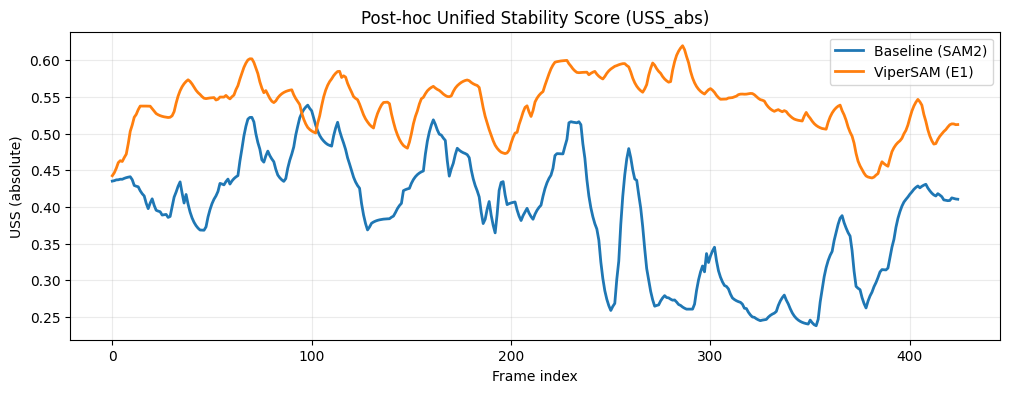}
    \caption{Post-hoc unified stability score (USS\textsubscript{abs}) for the desert car sequence.}
    \label{fig:desert_uss}
\end{figure}

\subsubsection{Jumping Lamb Sequence (322 frames)}

\begin{figure}[t]
    \centering
    \includegraphics[width=\columnwidth]{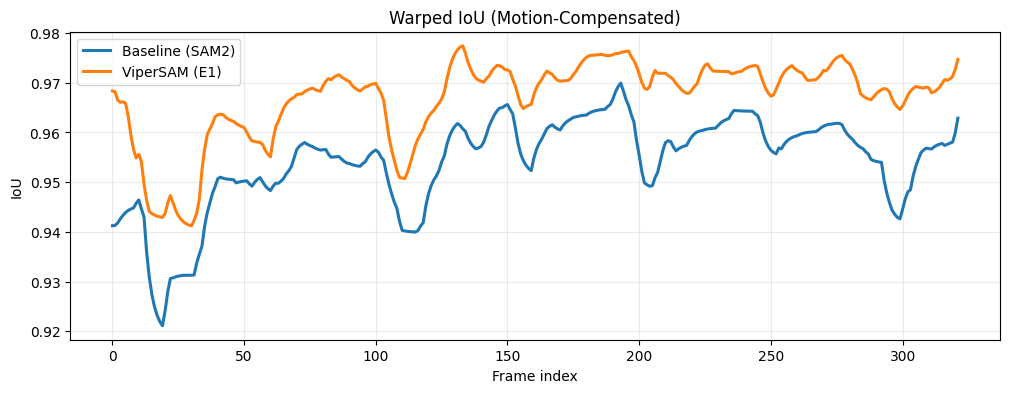}
    \caption{Warped IoU (motion-compensated) over time for the jumping lamb sequence.}
    \label{fig:lamb_wiou}
\end{figure}

\begin{figure}[t]
    \centering
    \includegraphics[width=\columnwidth]{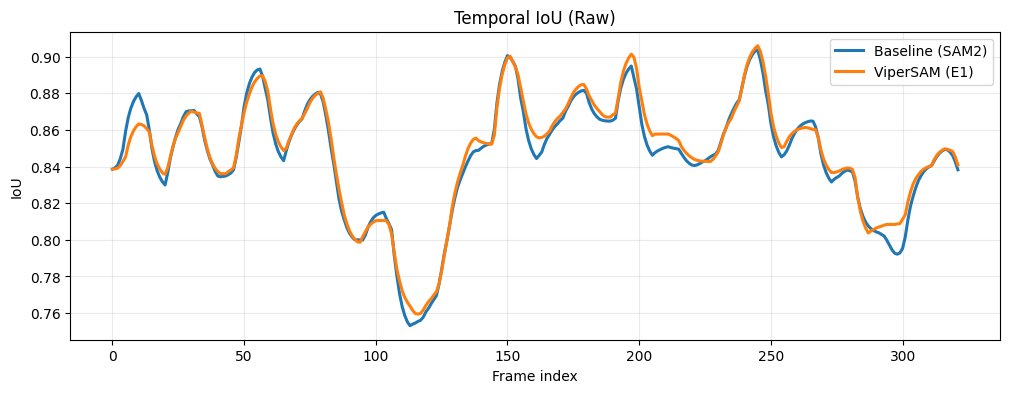}
    \caption{Temporal IoU (raw) over time for the jumping lamb sequence.}
    \label{fig:lamb_tiou}
\end{figure}

\begin{figure}[t]
    \centering
    \includegraphics[width=\columnwidth]{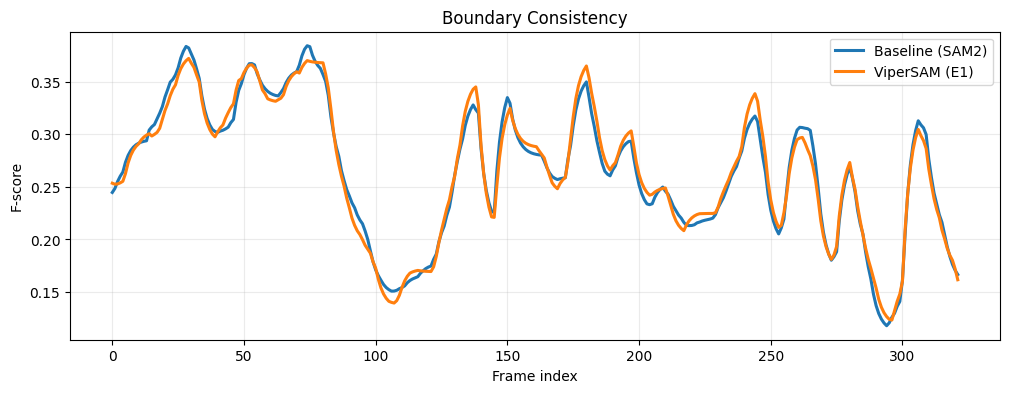}
    \caption{Boundary consistency (F-score) over time for the jumping lamb sequence.}
    \label{fig:lamb_boundary}
\end{figure}

\begin{figure}[t]
    \centering
    \includegraphics[width=\columnwidth]{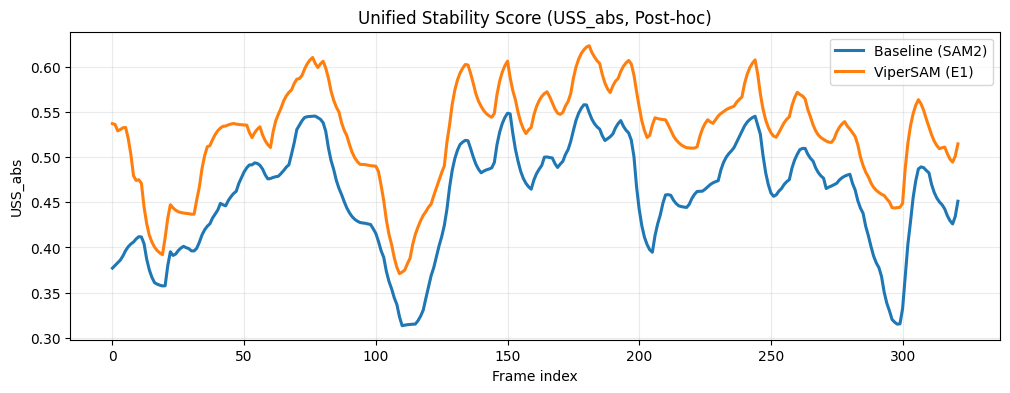}
    \caption{Post-hoc unified stability score (USS\textsubscript{abs}) for the jumping lamb sequence.}
    \label{fig:lamb_uss}
\end{figure}

\clearpage

\begin{strip}

\subsection{Qualitative Analysis and Detailed Visual Comparisons}
\label{sec:appendix_qualitative}

This subsection presents a comprehensive qualitative comparison between the baseline SAM2 and the proposed ViperSAM on the desert vehicle sequence, highlighting frame-by-frame segmentation behavior and improved temporal stability under weak prompt conditions.

\vspace{0.8em}

\centering
\includegraphics[height=0.78\textheight, keepaspectratio]{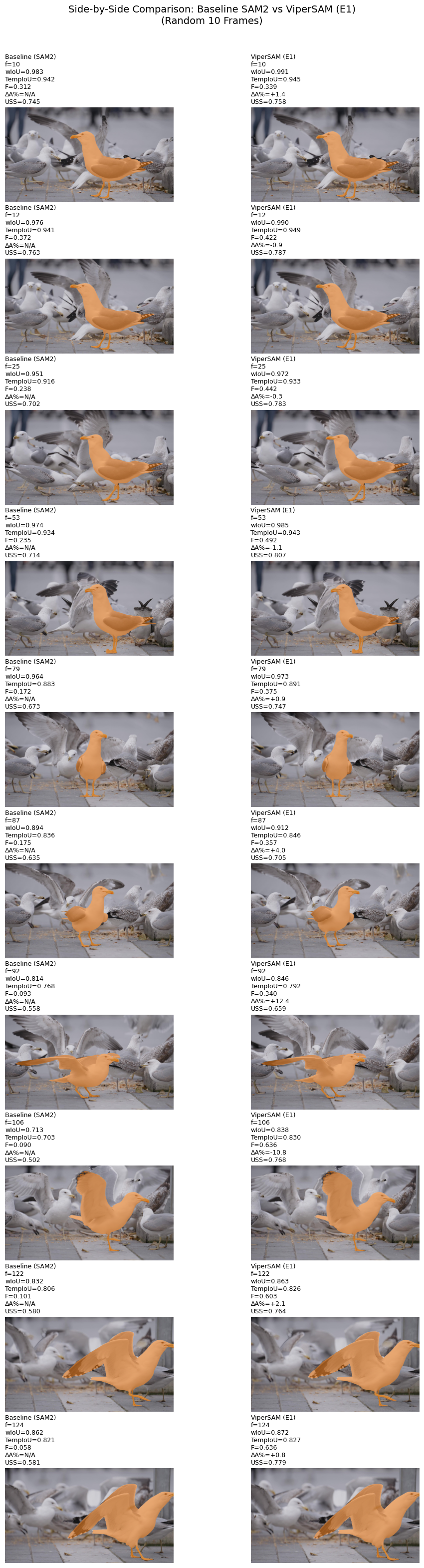}
\captionof{figure}{Comprehensive frame-by-frame comparison between baseline SAM2 and ViperSAM on the desert vehicle sequence. Each row corresponds to the same video frame, demonstrating improved temporal coherence and reduced mask flickering achieved by ViperSAM.}
\label{fig:appendix_bird_viper}

\end{strip}

\FloatBarrier

\clearpage

\bibliographystyle{ACM-Reference-Format}
\bibliography{references}

@inproceedings{kirillov2023segment,
  title     = {Segment Anything},
  author    = {Kirillov, Alexander and Mintun, Eric and Ravi, Nikhila and Mao, Hanzi and Rolland, Chloe and Gustafson, Laura and Xiao, Tete and Whitehead, Spencer and Berg, Alexander C. and Lo, Wan-Yen and Doll{\'a}r, Piotr and Girshick, Ross},
  booktitle = {Proceedings of the IEEE/CVF International Conference on Computer Vision (ICCV)},
  year      = {2023}
}

@article{ravi2024sam2,
  title   = {{SAM 2}: Segment Anything in Images and Videos},
  author  = {Ravi, Nikhila and Gabeur, Valentin and Hu, Yuan-Ting and Hu, Ronghang and Ryali, Chaitanya and Ma, Tengyu and Khedr, Haitham and R{\"a}dle, Roman and Rolland, Chloe and Gustafson, Laura and Mintun, Eric and Pan, Junting and Alwala, Kalyan Vasudev and Carion, Nicolas and Wu, Chao-Yuan and Girshick, Ross and Doll{\'a}r, Piotr and Feichtenhofer, Christoph},
  journal = {arXiv preprint arXiv:2408.00714},
  year    = {2024}
}

@article{ye2025entitysam,
  title   = {Zero-shot entity-level video segmentation with promptable foundation models},
  author  = {Ye, Anonymous and others},
  journal = {arXiv preprint},
  year    = {2025}
}

@inproceedings{perazzi2016benchmark,
  title     = {A Benchmark Dataset and Evaluation Methodology for Video Object Segmentation},
  author    = {Perazzi, Federico and Pont-Tuset, Jordi and McWilliams, Brian and Van Gool, Luc and Gross, Markus and Sorkine-Hornung, Alexander},
  booktitle = {Proceedings of the IEEE Conference on Computer Vision and Pattern Recognition (CVPR)},
  year      = {2016}
}

@inproceedings{caelles2017osvos,
  title     = {One-Shot Video Object Segmentation},
  author    = {Caelles, Sergi and Maninis, Kevis-Kokitsi and Pont-Tuset, Jordi and Leal-Taix{\'e}, Laura and Cremers, Daniel and Van Gool, Luc},
  booktitle = {Proceedings of the IEEE Conference on Computer Vision and Pattern Recognition (CVPR)},
  year      = {2017}
}

@inproceedings{voigtlaender2017online,
  title     = {Online Adaptation of Convolutional Neural Networks for Video Object Segmentation},
  author    = {Voigtlaender, Paul and Leibe, Bastian},
  booktitle = {British Machine Vision Conference (BMVC)},
  year      = {2017}
}

@inproceedings{wang2019video,
  title     = {Video Object Segmentation using Space-Time Memory Networks},
  author    = {Wang, Qiang and Zhang, Li and Bertinetto, Luca and Hu, Weiming and Torr, Philip H. S.},
  booktitle = {Proceedings of the IEEE/CVF International Conference on Computer Vision (ICCV)},
  year      = {2019}
}

@inproceedings{oh2019stm,
  title     = {Video Object Segmentation using Space-Time Memory Networks},
  author    = {Oh, Seoung Wug and Lee, Joon-Young and Xu, Ning and Kim, Seon Joo},
  booktitle = {Proceedings of the IEEE/CVF International Conference on Computer Vision (ICCV)},
  year      = {2019}
}

@inproceedings{brox2004high,
  title     = {High Accuracy Optical Flow Estimation Based on a Theory for Warping},
  author    = {Brox, Thomas and Bruhn, Andr{\'e}s and Papenberg, Nils and Weickert, Joachim},
  booktitle = {European Conference on Computer Vision (ECCV)},
  year      = {2004}
}

@inproceedings{sevilla2016optical,
  title     = {Optical Flow with Semantic Segmentation and Localized Layers},
  author    = {Sevilla-Lara, Laura and Sun, Deqing and Jampani, Varun and Black, Michael J.},
  booktitle = {Proceedings of the IEEE Conference on Computer Vision and Pattern Recognition (CVPR)},
  year      = {2016}
}

@inproceedings{jampani2017video,
  title     = {Video Propagation Networks},
  author    = {Jampani, Varun and Gadde, Raghudeep and Gehler, Peter V.},
  booktitle = {Proceedings of the IEEE Conference on Computer Vision and Pattern Recognition (CVPR)},
  year      = {2017}
}

@inproceedings{farneback2003two,
  title     = {Two-Frame Motion Estimation Based on Polynomial Expansion},
  author    = {Farneb{\"a}ck, Gunnar},
  booktitle = {Scandinavian Conference on Image Analysis (SCIA)},
  year      = {2003}
}

\end{document}